\documentclass[pdflatex,sn-mathphys-num]{sn-jnl}

\usepackage{graphicx}%
\usepackage{multirow}%
\usepackage{amsmath,amssymb,amsfonts}%
\usepackage{amsthm}%
\usepackage{mathrsfs}%
\usepackage[title]{appendix}%
\usepackage{xcolor}%
\usepackage{textcomp}%
\usepackage{manyfoot}%
\usepackage{booktabs}%
\usepackage{algorithm}%
\usepackage{algorithmicx}%
\usepackage{algpseudocode}%
\usepackage{listings}%

\usepackage{color}

\theoremstyle{thmstyleone}%
\theoremstyle{thmstyletwo}%

\theoremstyle{thmstylethree}%
\begin{document}

\title[Article Title]{Accelerating Quasi-Static Time Series Simulations with Foundation Models}

\author*[1]{\fnm{Alban} \sur{Puech}}
\equalcont{}

\author[2]{\fnm{François} \sur{Mirallès}}
\vspace{-1cm}
\equalcont{Equal contributors.}

\author[1]{\fnm{Jonas} \sur{Weiss}}
\author[2]{\fnm{Vincent} \sur{Mai}}
\author[2]{\fnm{Alexandre} \sur{Blondin Massé}}
\author[2]{\fnm{Martin} \spfx{de} \sur{Montigny}}
\author[1]{\fnm{Thomas} \sur{Brunschwiler}}
\author[3]{\fnm{Hendrik} \sur{F. Hamann}}

\affil[*]{Corresponding author: alban.puech1@ibm.com}

\affil[1]{\orgname{IBM Research - Europe, Zurich, Switzerland}}
\affil[2]{\orgname{Hydro-Québec Research Center, Quebec, Canada}}
\affil[3]{\orgname{IBM Research, NY, USA}}

\abstract{
Quasi-static time series (QSTS) simulations have great potential for evaluating the grid's ability to accommodate the large-scale integration of distributed energy resources. 
However, as grids expand and operate closer to their limits, iterative power flow solvers, central to QSTS simulations, become computationally prohibitive and face increasing convergence issues. 
Neural power flow solvers provide a promising alternative, speeding up power flow computations by 3 to 4 orders of magnitude, though they are costly to train.
In this paper, we envision how recently introduced grid foundation models could improve the economic viability of neural power flow solvers. Conceptually, these models amortize training costs by serving as a foundation for a range of grid operation and planning tasks beyond power flow solving, with only minimal fine-tuning required. 
We call for collaboration between the AI and power grid communities to develop and open-source these models, enabling all operators, even those with limited resources, to benefit from AI without building solutions from scratch.
}




\maketitle

\section{Introduction}\label{sec1}

The integration of an increasing number of distributed energy resources (DERs) like solar and wind in the distribution and transmission system, along with a higher penetration of electric vehicles (EVs), introduces significant uncertainty and variability in the planning and operation of electrical networks~\cite{IEA2023}. This is due to the compounded effects of a higher load, more dynamics in generation and load, and limitations in analytical capabilities to model these new dynamics~\cite{doe2020advancedtransmission, doe2022cleanelectricity, DOE2024Rep}.%

In this context, studying a few extreme case scenarios, like peak load, to characterize the electrical network is no longer sufficient. Instead, more comprehensive studies spanning full-year cycles, with fine temporal resolution and a broader range of load and generation conditions, have become necessary. Quasi-static time series (QSTS) simulation methodologies have first been successfully applied to carry out such studies on the distribution system, for example, to evaluate the impact of photovoltaic generation on the feeders~\cite{Azzolini2021}. As outlined in~\cite{supreme_oserplanning_2023,supreme_quasi-static_2024}, these approaches have been further enhanced by integrating an additional layer to model a complex transmission system and its operational constraints in more detail. QSTS simulations solve numerous successive power flows, related by discrete control actions. The process of identifying feasible control trajectories through a series of power flow solutions is both time-consuming for analysts and computationally intensive, as it involves sequential decision-making within complex system dynamics. For large grids, the complexity of this method rises sharply, as iterative power flow solvers' computation time scales quadratically with grid buses~\cite{Lagace2008}, which will make them impractical for future grids. Thus, adopting faster, more efficient computational approaches~\cite{DOE2024Rep} becomes critical. 

Recently proposed power grid foundation models (GridFMs)~\cite{puech2024optimal, hamann2024perspectivefoundationmodelselectric} present a promising approach. These data-driven emulators of power grids are trained in a self-supervised manner to approximate grid behavior. Unlike specialized AI models, GridFMs have the potential to be fine-tuned for a wide range of downstream tasks and grid configurations. Notably, they are expected to offer 3 to 4 orders of magnitude speedups in power flow solving~\cite{hamann2024perspectivefoundationmodelselectric}. In this work, we thus propose using a GridFM as a fundamental component of the QSTS simulation framework.

\section{QSTS Simulations - Operational Relevance and Obstacles to Larger Scale Application}\label{sec2}

\subsection{Operational Relevance}

QSTS simulations involve a sequence of steady-state power flows, conducted over time intervals ranging from a second to an hour. They account for changes in discrete controls of each equipment, time series profiles of loads, generation, interconnections with neighboring systems, and characteristics of energy storage systems and DERs. Operational guidelines and constraints provided by network planners are integrated into the analysis. For example, in~\cite{supreme_oserplanning_2023}, a QSTS simulation involves a virtual operator module that mimics driving actions. The primary goal of QSTS simulations is to identify potential stress points within prospective networks. As noted in~\cite{Deboever2017}, electrical networks are made of many interacting discrete control elements. Controllers' logic includes dead bands and delays, which make the systems' state time-dependent. As such, electrical networks are complex systems made of many interacting parts and can exhibit potentially undesirable emergent behavior. QSTS simulations include this notion of time dependence between successive power flows and discrete controls. Therefore, QSTS simulations facilitate more effective planning for network expansion and a better understanding of network operational limits. 

\subsection{Obstacles to Larger Scale Application}

Brute-force QSTS simulations require solving millions of power flows. To speed up these simulations, two main strategies exist: reducing the number of power flows, and accelerating their resolution. Reference~\cite{reno2018} illustrates the first strategy using an algorithm with varying time-step lengths. Reference~\cite{Deboever2018} adopts a different approach by recognizing that some power flow solutions can be reused due to similar inputs, leading to the development of an archive and retrieval system to avoid redundant computations. However, these techniques add complexity to the brute-force baseline, potentially making the QSTS solution sensitive to unforeseen use cases. The behavior of these algorithmic refinements in larger, more complex networks remains untested. Additionally, as noted by~\cite{Deboever2017}, performance gains depend on software module integration in QSTS simulation solutions. For instance, using external libraries to call power flow solvers and retrieving numerous parameters can also affect performance.

As the goal of QSTS simulations is to find potentially difficult operating conditions for a prospective network, classical iterative power flow solvers can have difficulties to converge~\cite{Tostado2019}. Although robust techniques have been developed, they often come at additional computational costs over the standard Newton-Raphson method. In practice, in case of difficult convergence, different solvers and parameters are tested through a trial-and-error process, which makes the overall QSTS simulation analysis time-consuming. As more DERs will be integrated into the grid in the future, and as grids become larger, these convergence issues are expected to occur more often~\cite{illconditioned}.

\section{AI Foundation Models to Overcome QSTS Computational Bottlenecks}\label{sec3}

In the previous section, we have identified two major computational bottlenecks of power flow solvers when used in QSTS simulations. To address them, we identify AI as a means of mitigation. Recent research results showed that AI, particularly graph neural networks (GNNs), could be capable \textit{neural} power flow solvers~\cite{DONON2020106547, varbella2024powergraphpowergridbenchmark, Tuo2023GraphNN, Boettcher}. 

\subsection{Shifting Computational and Numerical Stability Challenges from Runtime to Training}

Unlike traditional solvers, which iteratively solve non-linear equations, neural power flow solvers directly map inputs into the solution space, which particularly for large problems promise to be computationally more efficient, i.e. faster and overcoming convergence issues of iterative approaches. However, to gain this capability, AI models need to be extensively trained on large amounts of solved power-flow problems. In either case, significant computing power will have to be invested. With AI, however, this only needs to be done once during the training phase, relieving computational burdens at runtime. AI thus addresses the two key computational bottlenecks of iterative methods: their computational cost at runtime, and their numerical stability.

First, we have seen that the \textbf{computational cost} of iterative methods scales quadratically with the number of buses~\cite{Lagace2008}, while GNN-based neural power flow solvers have demonstrated linear complexity with the number of buses during inference~\cite{Wu_2021}. 
As a result, speedups of 3 to 4 orders of magnitude over conventional solvers have been demonstrated on IEEE test cases with up to 118 buses~\cite{DONON2020106547}. For larger grids, we thus expect even greater speedups.
%

Second, \textbf{numerical instabilities}, originating from ill-conditioned power flow cases~\cite{4682629, illconditioned} may lead to convergence challenges for iterative power flow solvers. With the direct mapping from input to output of neural power flow solvers, without iterative approximation of a system of non-linear equations, these challenges do not apply.
While more robust iterative power flow solvers may be capable of handling such ill-conditioned cases, their higher computational complexity~\cite{illconditioned} would make QSTS simulations intractable if used by default. 
Nevertheless, such solvers may still be valuable in solving ill-conditioned cases to augment the training dataset for neural power flow solvers.
This 
facilitates smooth system interaction of QSTS analysts, relieving them, among others, from having to experiment with different iterative solvers and their parameters when facing convergence issues~\cite{Xie2013}.

If need be, such AI solutions can be simply verified with the power flow equations, opening up the potential for their use in operations for certain applications.

\subsection{Making AI Models Economically Scalable using Foundation Models}

Iterative solvers approximate physics-based equations, and may thus in theory deal with arbitrary grid topologies and device parameters, i.e. cover an almost infinite solution space. It has been shown above, however, that this comes at a cost, and it does so for every single solution that needs to be computed. Neural power flow solvers, on the other hand, learn to map inputs to outputs based on training data. AI models, such as AI emulators for weather, have shown to be very efficient at inference time and are capable of generalizing to unseen cases or topologies~\cite{hakim2024dynamical}. However, this is always within limits and requires training on large datasets.
This training incurs large costs, since training data needs to be synthetically generated using iterative solvers and robust solvers for ill-conditioned scenarios. Nevertheless, these costs only occur once as non-recurring engineering (NRE) costs. Consequently, if such a neural power flow solver can create large numbers of solutions, these NRE costs may be easily amortized, as expected when used in the context of QSTS analysis, where different topologies are studied with many different input profiles (generation, load, and interconnections).

To further improve the cost-effectiveness of neural power flow solvers, we aim to exploit the concept of foundation models (FMs). These are comparably large models, trained in an unsupervised fashion on very large datasets without annotations, on specially designed pre-training reconstruction tasks~\cite{bommasani2022opportunitiesrisksfoundationmodels}. Once trained, they can relatively easily be modified for different applications or tasks within related fields~\cite{Wang_2023_CVPR}, and only need to be fine-tuned with comparatively little application-specific data. This prevents having to train many stand-alone, complex, application-specific AI models from scratch, with associated computation costs. FMs for power grids, GridFMs, may be well suited to solve the power flow problem, but also have the prospect to be capable of solving optimal power flow, contingency analysis or, as mentioned, solve grid expansion problems~\cite{hamann2024perspectivefoundationmodelselectric, puech2024optimal}.
There is also the prospect that in the long-term, such GridFMs may provide the means to indirectly overcome data silos and data regulation challenges, by providing pre-trained GridFMs as open-source assets, without having to explicitly share restricted and proprietary data repositories. Overall, this may enable operators with more limited budgets to exploit the most recent solver advancements and move towards more optimal and sustainable grid operations~\cite{DOE2024Rep}.


\subsection{Foundation Models - \textit{One Foundation for All and All Data for One}}
\begin{figure}
    \centering
\includegraphics[width=\linewidth]{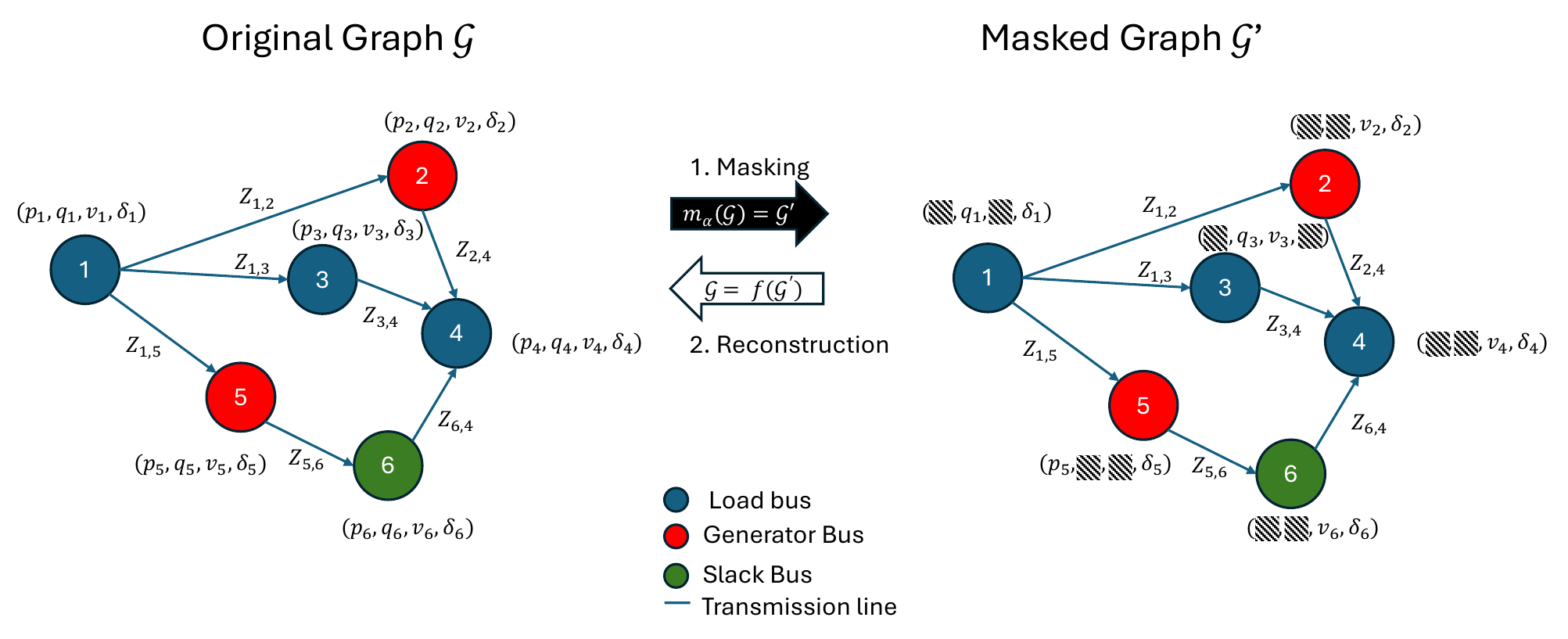}
    \caption{Masking and reconstruction steps for grid FM training. \textbf{1. Masking:} Given a graph representation of the transmission grid $\mathcal{G}$, the function $m_\alpha$ randomly masks node variables (independently of the bus type) with masking probability $\alpha$. The resulting masked graph is $\mathcal{G}'$. \textbf{2.  Reconstruction:} We assume the existence of a function $f$ that, given a masked graph $\mathcal{G}'$, returns the original graph $\mathcal{G}$. Our model is trained to approximate $f$.
    \label{fig:mae}}
\end{figure}

The vision for GridFM is to build \textbf{\textit{one foundation for all}}, or at least for as many downstream applications as possible. 
The choice of the pre-training task defines the physics that the model will learn to approximate, and determines the attainable downstream applications.
Since solving power flow equations is central to numerous applications in grid operations and planning, power flow reconstruction has recently been identified as a highly relevant task for pre-training a GridFM, which is discussed in length in~\cite{hamann2024perspectivefoundationmodelselectric}. In this approach, detailed in~\cite{hamann2024perspectivefoundationmodelselectric,puech2024optimal}, power flow data is represented as a graph, where nodes represent buses, and node features \( (p_i, q_i, v_i, \delta_i) \), correspond to active power, reactive power, voltage magnitude, and voltage angle, respectively. Edges correspond to transmission lines or transformers. We envision to train an autoencoder, such as GraphMAE~\cite{graphmae, graphmae2}, to reconstruct randomly masked node features, as shown in Figure \ref{fig:mae}. This corresponds to solving the power flow equations. 

Previous neural power flow solvers, such as~\cite{varbella2024powergraphpowergridbenchmark}, trained separate models for each grid topology and set of parameters. 
Training a GridFM, however, capable of generalizing across different grid topologies, parameters, and load scenarios, will require \textbf{\textit{all data for one foundation}}. 
Thus, a large and diverse collection of solved power flow problems will be required to train a single foundation model. 
When trained this way, these models generalize well to graphs unseen during training \cite{mao2024positiongraphfoundationmodels,liu2024graphfoundationmodelssurvey}, and specifically also for power grid tasks~\cite{VarbellaCFS2023}. In the latter, a model was trained on different grids for detecting cascading failures, and was tested on yet another grid, unseen during training, with an impressive 96\% accuracy.

In~\cite{gnnreviewpowergrid}, the authors propose a dataset containing power flow solutions under real load conditions with 15-minute resolution over a year, across four different grid topologies. Other work has generated power flow cases using random perturbations of the load at each node, as well as by introducing topology variations such as dropping lines or generators at random~\cite{lovett2024opfdatalargescaledatasetsac,piloto2024canosfastscalableneural}.  
To further enrich these datasets, we plan to incorporate more grid topologies from PGLIB-OPF~\cite{pglib} and generate additional power flow cases using different solvers from the PandaPower library~\cite{pandapower}. Particular attention should be given to cases that lead to low solution accuracy or convergence difficulties with iterative solvers.
Another focus will be on generating cases that reflect real-world industrial scenarios through realistic load, topology perturbations, and control strategies. These should capture the statistical distribution of cases typically encountered during tasks such as QSTS simulations, while also meeting the operational constraints of electrical utilities. 
This will require collaboration between AI and power experts.

\section{Conclusion and Future Outlook}

QSTS simulations demonstrated their potential for assessing the grid's ability to handle the large-scale integration of distributed energy resources. These simulations require solving numerous power flows. However, as grids grow larger and are operated closer to their limits, iterative power flow solvers become too computationally expensive and increasingly face convergence issues. Neural power flow solvers offer a solution by learning a direct mapping between inputs and solutions, avoiding iterative numerical approximations that may lead to instability. As these AI models shift computational efforts away from runtime to training, they have the promise to accelerate power flow resolution by 3 to 4 orders of magnitude.

In this paper, we envision how recently introduced grid foundation models could make neural power flow solvers economically viable and speed up QSTS simulations. These models could potentially be applied to a wide range of tasks beyond solving power flow, by minimal fine-tuning of the base model. We thus expect them to create significant value for operation and planning, beyond QSTS simulations.

We thus call for the AI and power grid communities to join forces into a common goal of developing and open-sourcing such power grid foundation models. These models will enable all grid operators -- including those with limited resources such as data, R\&D teams, or funding -- to benefit from the advantages of AI, without having to build their own AI solutions from the ground up.

\clearpage

\section*{Author contributions}
\begin{itemize}
    \item A.P. and F.M.: Conceptualization, Investigation, Writing – Original Draft
    \item J.W. and V.M.: Investigation, Writing – Original Draft
    \item A.B.M. and M.D.M.: Investigation, Writing – Review \& Editing
    \item T.B. and H.F.H.: Project Administration
\end{itemize}

\bibliography{bibliography}

\end{document}